\documentclass{article}

\usepackage{microtype}
\usepackage{graphicx}
\usepackage{subfigure}
\usepackage{booktabs} 
\usepackage{hyperref}

\usepackage{subcaption} 

\usepackage[accepted]{icml2024}

\usepackage{amsmath}
\usepackage{amssymb}
\usepackage{mathtools}
\usepackage{amsthm}

\usepackage[capitalize,noabbrev]{cleveref}

\theoremstyle{plain}

\theoremstyle{definition}

\theoremstyle{remark}

\usepackage[textsize=tiny]{todonotes}
\usepackage{todonotes}
\usepackage{graphicx} 
\usepackage{subcaption} 

\icmltitlerunning{Position: Cracking the Code of Cascading Disparity Towards Marginalized Communities}

\begin{document}

\twocolumn[
\icmltitle{Position: Cracking the Code of Cascading Disparity Towards Marginalized Communities}


\begin{icmlauthorlist}
\icmlauthor{Golnoosh Farnadi}{yyy}
\icmlauthor{Mohammad Havaei}{yyy}
\icmlauthor{Negar Rostamzadeh}{yyy}
\end{icmlauthorlist}

\icmlaffiliation{yyy}{Google Research, Montreal, Canada}

\icmlcorrespondingauthor{Golnoosh Farnadi}{gfarnadi@google.com}

\icmlkeywords{Responsible AI, Machine Learning}

\vskip 0.3in
]



\printAffiliationsAndNotice{} 

\begin{abstract}


The rise of foundation models holds immense promise for advancing AI, but this progress may amplify existing risks and inequalities, leaving marginalized communities behind. In this position paper, we discuss that disparities towards marginalized communities – performance, representation, privacy, robustness, interpretability and safety – are not isolated concerns but rather interconnected elements of a \emph{cascading disparity phenomenon}. We contrast foundation models with traditional models and highlight the potential for exacerbated disparity against marginalized communities. Moreover, we emphasize the unique threat of cascading impacts in foundation models, where interconnected disparities can trigger long-lasting negative consequences, specifically to the people on the margin. We define marginalized communities within the machine learning context and explore the multifaceted nature of disparities. We analyze the sources of these disparities, tracing them from data creation, training and deployment procedures to highlight the complex technical and socio-technical landscape. To mitigate the pressing crisis, we conclude with a set of calls to action to mitigate disparity at its source. 

\end{abstract}

\section{Introduction}

Foundation models with their ability to learn and adapt across various domains, are rapidly transforming the landscape of AI. However, these large-scale models that often trained on massive, unfiltered datasets, pose various risks for marginalized communities. Foundation models can perpetuate and amplify existing biases, leading to disparities in performance, privacy, robustness, model understanding, and even the generation of harmful content for marginalized communities. For example, large language models~(LLMs) often perform worse on language tasks involving dialects spoken by low-resource languages~\cite{liang2021towards}. An image generation or diffusion model primarily associating certain professions with specific genders or ethnicities~\cite{luccioni2023stable}. Multimodal models can struggle with recognizing or classifying images of people from marginalized groups, particularly those with darker skin tones or features that do not align with dominant beauty standards~\cite{buolamwini2018gender,schwemmer2020diagnosing}. A voice assistant models often misinterpret commands spoken with a regional accent which leads to frustration and exclusion for the user~\cite{tatman2017gender}. Multimodal models might generate hateful or offensive content that perpetuate discrimination and incite violence against already vulnerable groups~\cite{zellers2019defending}.

While extensive research has focused on identifying and addressing specific types of disparities (see Section~\ref{sec:disparity_type}), a holistic understanding of how these disparities are interconnected remains largely unaddressed. This narrow focus can worsen existing biases and introduce new ones, disproportionately harming marginalized communities. For example, counterfactual data augmentation~\cite{gardner2020evaluating} that have shown promising out-of-domain generalizability~\cite{samory2021call}, if implemented without careful consideration, can reinforce harmful stereotypes or lead models to violate established social norms~\cite{sen2022counterfactually}.

Moreover, standard evaluation benchmarks, such as StereoSet~\cite{nadeem2020stereoset}, often focus on surface-level assessments of foundation models using non-robust prompting metrics or post-deployment downstream task evaluations. These approaches fail to directly probe the deeper sources of bias embedded within the models.

In this position paper, we argue that these disparities are interconnected elements of a \textbf{cascading disparity phenomenon} affecting marginalized communities. We discuss that representational disparities within the model lie at the root of this phenomenon. The distinct, complex distributions representing marginalized communities are often insufficiently captured during training, leading to the "flattening" of their representations within the model. This, in turn, manifests as performance disparities, reinforcement of stereotypes, privacy violations, and other types of harmful disparities. By analyzing the sources of disparities throughout the lifecycle of foundation models, from data collection and training to adaptation and deployment, we highlight the complex technical and socio-technical landscape that shapes this problem. We urge researchers to move beyond conventional loss function optimization when training foundation models. It's vital to develop metrics that directly assess the quality of representations with regards to marginalized communities, ensuring that the model learns the nuanced, low-dimensional manifolds associated with these groups.  Additionally, we advocate for investigating how a model's capacity should be dynamically allocated across different distributions to achieve a more equitable representations. Our contributions in this position paper are as follows: 1) Identifying and categorizing disparities in foundation models that disproportionately impact marginalized communities (Section~\ref{sec:disparity_type}). 2) Defining the cascading disparity phenomenon and how it stems from representational disparities (Section~\ref{sec:cascading}). 3) Analyzing the origins of disparities across the lifecycle of foundation models (Section~\ref{sec:disparity_source}). 4)Providing a list of call to actions to address representational disparities in foundation models (Section~\ref{sec:call_for_action}).


\section{Marginalized Communities}
\label{sec:marginalized}

Before discussing how and why the current way of training and deploying models creates disparity towards marginalized communities, it's important to clearly define what we mean by marginalized communities or ``data at the margin'' in the context of machine learning~(ML). Marginalized communities \cite{allman2013sociology} or as Williams and White put it ``marginalized from mainstream society''~\cite{williams2003conceptualising} refers to groups systematically excluded and discriminated against based on factors like race, ethnicity, gender, sexual orientation, socioeconomic status, disability, religion, or other identity aspects \cite{allman2013sociology, williams2003conceptualising}. This historical exclusion often results in ongoing lack of representation, resources, and ongoing discrimination.
Such historical exclusion is reflected and amplified in ML applications. Data from marginalized communities is often missing, underrepresented, or misrepresented in training datasets. This leads to several data-centric challenges for ML models: 1) \textbf{Small sample size:} Marginalized communities are underrepresented in datasets. 2) \textbf{Disparate distribution:} The data distributions associated with marginalized communities may differ significantly from the majority population. This can encompass factors like demographics, language use, or behavioral patterns. 
3) \textbf{Complex distributions:} Data may exhibit nuances and complexities due to intra-group diversity, cultural patterns, or unique historical contexts.

These factors create challenges for ML models, making it difficult to represent marginalized communities accurately.  In statistical terms, these low-sample classes form the ``long tail'' of distributions. While this typically refers to low-occurrence events, in this context, it highlights data samples rarely seen during training, despite their real-world prevalence not necessarily being lower than more common data classes. 

Note that this definition focuses primarily on the data-centric aspects of marginalization in machine learning
and it’s essential to acknowledge that definitions of marginalized communities are not solely technical but also inherently social and political. In the context of technology and socio-technical systems, marginalized communities can be understood as groups of people who experience: i) \textit{Systemic Disadvantage}: These communities face historical, social, political, and economic barriers that limit their access to opportunities, resources, and power. This systemic disadvantage often stems from factors like discrimination, prejudice, and social exclusion.  ii) \textit{Data Exclusion and Invisibility}: Marginalized communities may be underrepresented or even invisible within data sets used to train and develop technological systems. And iii) \textit{Limited Agency and Participation}: Marginalized communities may have limited opportunities to participate in the design, development, and deployment of technological systems that significantly impact their lives. 

\section{Types of Disparity}
\label{sec:disparity_type}

In this section, we present how foundation models systematically disadvantage marginalized communities through the following eight disparities.

\noindent \textbf{Embedding/Representation disparities} Foundation models serve as powerful tools for representation learning with the goal of automatically capturing meaningful and generalized features from the data that makes it easier to extract useful information in down stream tasks. A good representation is one that captures the underlying explanatory factors for the observed input. As such representation learning is closely linked to manifold learning with the hypothesis that high dimensional data lies on a low dimensional manifold~\cite{gorban2018blessing}. It is generally understood that learning complex features leads to better generalization in downstream tasks~\cite{bengio2013representation,natekar2020representation}. A representation should be expressive enough to capture complexities expressed through factors of variation for every subgroup in of the input space. With that notion, representation disparity is defined as constrained complexity in learned embeddings due to data limitations, resulting in challenges in manifold creation. Prior work has either focused on removing sensitive attributes with adverserial debiasing~\cite{zhang2018mitigating} and contrastive learning~\cite{tian2020makes} or maintaining semantic distances in the embedding space~\cite{zafar2017fairness,beutel2017data, zhang2018mitigating}, addressing the limited complexity issue. 

\noindent \textbf{Performance disparities} Performance disparity is defined as disparities in model performance between majority and minority populations in downstream tasks. Previous work has shown such performance gap is manifested in health-care~\cite{hall2023vision}, text summarization~\cite{yang2023exploring}, translation~\cite{prates2020assessing}, image classification~\cite{ali2023evaluating}, and recommender systems~\cite{moradi2023tidying}. Extensive research has explored the performance gap, with various fairness metrics like demographic parity, accuracy parity, equal opportunity, and equalized odds~\cite{hardt2016equality} reflecting these disparities for marginalized communities.

\noindent \textbf{Privacy disparities} Privacy disparity is defined with a propensity for memorization more pronounced for marginalized communities~\cite{carlini2022quantifying}. Limited model capacity results in prioritized generalization for larger populations, exacerbating privacy concerns~\cite{tramer2016stealing, carlini2018audio} and catastrophic forgetting~\cite{luo2023empirical}. Existing work have shown privacy-enhancing technologies such as differential privacy in SGD~(DP-SGD)~\cite{abadi2016deep} that rely on gradient clipping and noise injection,  disproportionately degrade accuracy of marginalized communities~\cite{bagdasaryan2019differential,malekmohammadi2024mitigating}. Furthermore, model compression techniques on foundation models such as iterative magnitude pruning~\cite{maene2021towards}, which can result in enhancing overall privacy of the model , proportionately impact on the accuracy of communities at the margin~\cite{hooker2020characterising,hooker2019compressed,tran2022pruning,hashemizadeh2023balancing}

\noindent \textbf{Robustness disparities} Robustness disparity is the variation of the performance, accuracy and reliability of the ML model across different populations that could particularly impact the marginalized communities. This issue arises from inadequate representation of these communities in various stages of ML development, including (i) data creation (ii) model development and (iii) deployment. On \textit{data} side, marginalized communities are usually out-of-distribution samples. During the learning, their distribution is often miss-represented and under specified. This would make marginalized data also more prone to adversarial and poisoning attacks due to the lack of generalization of the model for these groups \cite{madry2017towards, athalye2018obfuscated,ma2022tradeoff}. Finally during the deployment process, conditions that the model is deployed for the marginalized communities are often dismissed and they lack proper testing for edge cases, group-specific perturbations and robustness testing towards factors of variation within these groups, that leads to increased vulnerability to adversarial attacks and failures in marginalized populations.

\noindent \textbf{Hallucination\footnote{Also referred to as ``Confabulation'' in literature} disparities}. It is widely acknowledged that Large Language Models (LLMs) and by extension Visual Language Models (VLMs) suffer from Hallucinations. These instances manifest as the model confidently generates output that while seeming plausible, are unreasonable or factually untrue with respect to the source of information. While the source of hallucination is not yet fully understood, hallucination in LLMs typically arise from the inherent data limitations in the training data and complexity of the model architecture~\cite{ji2023survey,dziri2022origin}. Hallucination disparity is defined as an elevated likelihood of generating fabricated or hallucinated outputs for marginalized communities due to data limitations. ~\citet{wang2020exposure} showed that hallucinations are more prevalent for out-of-domain distributions compared to in-domain distributions for Neural Machine Translation. \citet{cohen2018distribution} also demonstrated that a mismatch in distribution between source and target domains in image translation causes the model to hallucinate confounding factors when generating samples from the target domain.

Insufficient information and memorization tendencies contribute to the model making erroneous assumptions about the data and consequently leading to higher likelihood of generating inaccurate outputs for marginalized communities~\cite{wang2020exposure,cohen2018distribution,arjovsky2019invariant, guo2018deep}. These erroneous assumptions and misrepresentations manifest themselves in the learned manifold of the foundational model. From this perspective, hallucination disparity is closely linked to representation disparity, wherein the model has not acquired expressive representations of marginalized groups. 
The model's failure to capture the subtleties inherent to these groups within its learned representations is a contributing factor to the emergence of hallucinations in its generated outputs.

Note that while hallucination gap can be categorized under the broader category of performance disparity, we believe highlighting hallucinations as a distinct issue allows us to emphasize the importance of addressing outputs that are factually incorrect or misleading. This has significant implications for model reliability and factfulness, that warrants focused attention beyond the performance disparity in downstream tasks that are often focused on supervised learning tasks with existing measures such as demographic parity or equalized odds. 

\noindent \textbf{Model Understanding disparities} Foundation models, with their vast number of parameters and complex training data, are often challenging to fully comprehend. The lack of explainability is amplified when the model's training data is not representative of diverse populations~\cite{du2020fairness}. Trying to understand a model that generates text with underrepresented groups might lead to inaccurate assumptions or explanations that are not truly reflective of how the model works. This can result in misinterpreting the outputs and assigning incorrect attributions to the model’s behavior. Since model decisions are based on countless data points and parameters, determining the specific reasons for a particular output is often difficult~\cite{zhao2023explainability}. This becomes even more challenging when a model has not been exposed to sufficient data or diverse perspectives regarding marginalized groups. Without deep knowledge of a model's inner workings and data, there is a risk of simplifying its behavior. Moreover, the people developing the models and interpreting will be less likely to identify issues related to marginalized groups. Such oversimplifications may overlook the nuances or complexities involved when the model interacts with topics related to marginalized communities.


\noindent \textbf{Model Multiplicity/Underspecification disparities} Due to model uncertainty, the predictions or text generated by the model can be seemingly arbitrary or random when addressing topics related to marginalized groups~\cite{black2022model}. The model might generate responses that are off-topic, insensitive, or even harmful. Foundation models often involve stochastic (random) processes during the generation of text. This randomness, combined with a lack of understanding of underrepresented groups, can amplify the arbitrary nature of the outputs, making them increasingly unpredictable~\cite{ganesh2023impact,d2022underspecification}. The stochastic elements in these models can sometimes exaggerate the biases present in the training data. This can lead to the generation of random outputs that inadvertently amplify stereotypes or misinformation concerning marginalized communities, Ganesh showed that due to model multiplicity, the random behavior of the model is higher for marginalized groups~\cite{ganesh2024empirical}.

We intend to underscore how model design limitations (e.g., architectures that are too broad or too narrow) can specifically lead to ambiguity and uncertainty in model behavior. This connects model architecture choices directly to issues of bias and fairness towards marginalized communities. Our classification system aims to draw attention to these specific nuances within the broader \emph{performance disparity} category. This will facilitate more targeted analysis of existing mitigation strategies, even while acknowledging the interconnected nature of these issues. 

\noindent \textbf{Safety disparities} Regarding the detection and mitigation of safety concerns, 
conflicts could arise due to differing moral values and cultural contexts between groups~\cite{scherrer2023evaluating,benkler2023assessing}. Existing work show that current models may reflect dominant Western cultural biases and values~\cite{rao2023ethical}.  E.g.,  A model used to detect hate speech may have difficulty identifying slurs or harmful language directed towards certain groups due to underrepresentation in its data~\cite{deshpande2023toxicity}. Similarly, an AI model used to write text or stories may generate content that reinforces stereotypes or reflects biases against marginalized groups if it has limited exposure to inclusive data~\cite{blodgett2020language}.

\section{Cascading Disparity: A Systemic Issue of Interlinked Disparities due to Embedding Disparity}
\label{sec:cascading}

The eight disparities that we discussed in the previous section, while distinct in nature, are not independent issues. They interact and reinforce each other, creating a cumulative negative impact on marginalized communities. In this section, we show that at the root of this complex issue lies embedding disparity that its influence cascades exacerbating other forms of disparity.


Imbalances in how foundation models represent various groups within their embedding space directly contribute to performance disparities in downstream tasks. When models lack expressive representations of marginalized groups, their performance suffers in tasks that involve those groups. 
Such lack of proper manifold creation for marginalized communities and their underrepresentation, also force the model to use it's capacity to memorize specific data points instead of learning generalizable representations. Marginalized communities are often underrepresented which can make their data points seen as outliers, and make them more susceptible to privacy attacks under the privacy onion effect, outlined by Carlini et al.~\cite{carlini2022privacy}. If the model memorizes specific data points (common for marginalized groups), its behavior is inconsistent and challenging to even explain or interpret. 

Models with poor embedding representations of marginalized groups are also less robust, leading to higher uncertainty. Limited representation of marginalized groups in the embedding space leads to unpredictable and arbitrary predictions due to increased sensitivity to minor input changes. 
And while hallucinations can occur for various reasons, they often stem from model uncertainty. When a model is less certain about how to handle data from a marginalized group, it is more likely to fabricate or "hallucinate" details that are not grounded in the data.

The lack of manifold embedding disparity in foundation models is a critical safety concern. 
If marginalized groups are misrepresented in the embedding space, the model may fail to recognize their unique perspectives, cultural contexts, needs, and languages specific to those communities~\cite{jha2024beyond,qadri2023ai}. This lack of representation can result in outputs that overlook the concerns of marginalized groups or provide inaccurate or inappropriate responses to their needs, effectively excluding them from the benefits these models could offer or even perpetuate harmful stereotypes, reinforce exclusion, and amplify hate speech. When a model fails to learn the patterns associated with diverse perspectives, it struggles to generalize its knowledge to unseen scenarios or when presented with prompts related to underrepresented groups. This limitation results in unpredictable and uncertain outputs. Insufficient and imbalanced embedding space can cause the model to associate certain attributes more strongly with some groups than others. This might result in inconsistent outputs, where the same prompts produce different responses depending on the perceived identity of the subject. This uncertainty can lead to outputs that are biased or discriminatory.
    

We intentionally focused on eight well-established categories of responsible AI to demonstrate the interconnectedness of disparities. We highlight how overlooked intersections can magnify harms, alongside widening performance gaps. To mitigate the cascading negative impacts on marginalized communities, we need to address these interlinked disparities at the core. Next, we discuss the sources of manifold embedding disparities in foundation models.

\section{Sources of Disparity}
\label{sec:disparity_source}

While both traditional ML models and foundation models can exhibit disparities, the nature and sources of these disparities can differ significantly. Foundation models, due to their scale and complexity, present unique challenges in terms of disparity identification, mitigation, and societal impact. While the previous section highlighted the concerning disparities that can arise with foundation models, understanding the sources of these disparities is crucial for effectively addressing them. In this section, we explore how and why current practices in foundation model development can amplify and perpetuate harmful disparities, particularly against marginalized communities, and how they can differ from traditional models in several key ways.

\subsection{Design and Data Collection}
One fundamental challenge in training ML models, including traditional and foundation models, revolves around the obstacles posed by data. Here, we argue how and why data issues are contributing to representation disparity. Traditional ML models, typically require smaller, and domain-specific datasets tailored to the specific task, and the disparities often arise from the specific data used to train the model. If the data is imbalanced or contains inherent historical disparities, the model will likely learn and perpetuate those disparities. In foundation models, due to their massive scale and reliance on diverse datasets, they can be susceptible to a wider range of data biases. Data used to train foundation models can be orders of magnitude larger than those used in traditional ML, which presents challenges in data storage, processing, and ensuring data quality. Uneven or non-random sampling methods can lead to datasets that underrepresent or misrepresent certain demographics, creating skewed data distributions that disadvantage marginalized communities~\cite{passi2019problem}. The accessibility and availability of data can also vary across different groups~\cite{olteanu2019social}. Moreover, societal biases ingrained in cultural norms, and historical data can be inadvertently embedded within datasets, leading to models that reflect and amplify these biases~\cite{pedreschi2009measuring,richardson2019dirty}. Finally, differences in how individuals interact with technology, i.e., digital gap~\cite{hargittai2011minding}, or provide data~\cite{olteanu2019social} can introduce biases into datasets. For instance, marginalized communities may have limited access to technology or may be hesitant to provide data due to privacy concerns, leading to models that are less accurate or perform poorly on tasks involving these groups~\cite{molamohammadi2023unraveling}

\subsection{Training Procedure and Learning Algorithm}
\label{sec:learning_algorithm}



Disparities in traditional models are typically studied through the data or the chosen loss function. However, in foundation models, due to their complex learning processes and interactions with vast amounts of data, they can exhibit disparities that were not present in the training data or algorithms. These emergent disparities can be difficult to anticipate and address. Bellow, we discuss the source of these disparities during the training process that although they can occur in traditional models, the magnitude of the issue in foundation model can differ significantly. Sara Hooker~\cite{hooker2021moving} discussed how we should look beyond data to discuss disparities in ML models and consider the choices that we make, e.g., the algorithms or hyper-parameters, to study algorithmic discrimination.  Bellow, we extend her analysis and discuss the impact of various choices that we make during training that can have a significant impact on the outcome:

\noindent\textbf{Loss function}: Foundation Models diverge significantly from traditional ML  models in terms of their loss function design. Foundation models often employ self-supervised learning paradigms that emphasize learning the sequential structure of data instead of directly optimizing for performance on a specific task~\cite{brown2020language,devlin2018bert}. A common example of this is the "next token prediction" objective, where the model is trained to predict the next word or token in a given sequence.  This shift towards sequence learning has resulted in remarkable improvements in the language capabilities of large language models, among other areas. In traditional supervised ML, loss functions are designed to identify features and patterns in the data that are strongly correlated with a particular label or target variable. The model learns to prioritize the features most relevant to the task at hand. However, foundation models trained with sequence prediction objectives learn representations that capture the sequential dependencies and underlying structure of the data, irrespective of specific labels. This enhances their ability to generalize to a diverse range of downstream tasks without the need for extensive task-specific fine-tuning~\cite{radford2019language}. However, this focus on sequential learning has a profound impact on how foundation models process and learn from data, e.g., the pre-training data for the model might reflect biases related to the order or sequence in which information is presented such as sequential exposure to gendered language can reinforce gender stereotypes~\cite{bolukbasi2016man,sun2019mitigating}. Existing work indicates that in scenarios where there exists a non-linear relationship between group membership (e.g., considering demographics like race or gender) and a specific outcome, using a single linear classifier often leads to a performance trade-off. One, or perhaps both, of the groups involved will experience a decline in model performance~\cite{dwork2018decoupled}. This is because linear classifiers, by nature, struggle to capture the complexity of these non-linear relationships. In certain cases, however, incorporating information about the group differences directly into the design of the machine learning model can lead to the development of simpler learned functions (mathematical representations) that ultimately enhance the performance across various groups~\cite{dwork2018decoupled,suresh2019framework}. By understanding the nuances of the group differences, models can be tailored to better learn the diverse patterns within the data. However, such mitigation cannot simply be transferred to foundation models due to the complexity of identifying marginalized groups. 
    
\noindent\textbf{Aggregation Method}: The aggregation method for calculating the overall loss across samples influences how the model learns the representation of different groups within data. When data points are aggregated with similar weights, the model tends to prioritize learning the distribution of larger populations. If a loss function overemphasizes majority groups, existing work show that this leads to under-representation or misrepresentation of marginalized groups~\cite{suresh2019framework,mehrabi2021survey}. This results in the model neglecting the complexities of marginalized communities, often leading to performance gaps and a tendency towards memorization rather than generalization in representing these groups (as discussed in Section~\ref{sec:disparity_type}).
    
\noindent\textbf{Data order}: The order in which data points are presented during training, particularly in the context of marginalized groups, can significantly impact the model's performance~\cite{ganesh2023impact}. When data points are read randomly, there is a risk of catastrophic forgetting~\cite{kirkpatrick2017overcoming}, that the model will tend to forget the patterns associated with marginalized groups encountered earlier in the training process. This oversight often results in suboptimal model performance for these groups. However, research into the effects of fine-tuning has demonstrated that strategically positioning marginalized group data toward the end of the training process can improve their representation in the model~\cite{dodge2020fine}. This is because, in the fine-tuning stage, the model has a chance to reinforce its understanding of the marginalized group's patterns while the knowledge of the majority groups remains relatively stable, potentially leading to better performance for the marginalized communities.
    
\noindent\textbf{Batch size}: In the presence of random data selection, the aggregation of gradients from various data samples also plays a crucial role in determining the model's gradient norms and update directions. Due to the inherent distributional differences between marginalized communities and majority groups, gradient updates from these groups are likely to conflict, potentially in terms of both their directions and magnitudes~\cite{suresh2019framework}. When the training dataset comprises a significantly larger proportion of data from the majority groups, the model's overall learning trajectory is likely to be dominated by these majority data points. As a result, the learning of patterns from marginalized groups, effectively their 'voices', can be suppressed during the training process.

\noindent\textbf{Batch Composition}: Similar to the considerations of batch size and data presentation order, the composition of the training batch can also exert significant influence on the model's behavior, particularly when it comes to learning about marginalized communities. If smaller batch sizes are employed, and these batches disproportionately consist of data from a marginalized community, there is a higher likelihood that the aggregated gradient updates from these batches will notably influence the model's learning trajectory, making it more receptive to the patterns and characteristics present in the marginalized group's data.

\noindent\textbf{Learning Rate}: Learning rate and training duration exhibit a disproportionate influence on error rates on marginalized communities specifically those on the long tail of the distribution. Studies on deep neural network memorization demonstrate a delayed learning process for the marginalized communities~\cite{jiang2020characterizing}. Consequently, a common early stopping approach can carry the potential to systematically bias performance against certain data distributions~\cite{hooker2021moving}.



\subsection{Deployment and Adaptation}
One of the challenges of traditional supervised models are their adaptability to a new task or domain. Traditional ML models are often designed and trained from scratch for specific tasks. This focused training while beneficial for performance on targeted task, restricts their transferability to new tasks and domains.
Transfer learning approaches try to address this limitation by
leveraging knowledge from source tasks to target tasks and domains without the need to train from scratch. These approaches could be Homogeneous \cite{zhuang2020comprehensive,weiss2016survey}, when the feature and label spaces remain consistent across domains but differ slightly in their distributions. Homogeneous techniques aim to minimize these distribution discrepancies. On other other hand, transfer learning approaches could be Heterogeneous techniques~\cite{zhuang2020comprehensive,weiss2016survey} when labels, and feature spaces differs and they aim to bridge the gaps between different distributions and feature spaces.  

Creation of the foundation models created a significant boost to the improvement of transferablity of the knowledge between domains, and tasks. However, 
unlocking the potential of foundation models for real-world applications requires effective adaptation to specific downstream tasks. There are two primary adaptation paradigms for foundation models: \textbf{fine-tuning} and \textbf{prompt engineering}. In this section, we explore their distinct strengths, and limitations, and discuss how each approach can potentially mitigate or amplify the disparities towards marginalized community.

\subsubsection{Fine-tuning} 

One of the most widely used transfer learning approaches, that is usually a homogeneous approach, is fine-tuning~\cite{brown2020language}. 
Fine-tuning offers a means to tailor foundation models to specific tasks and domains by adjusting internal parameters based on task-specific data and domain knowledge. This can significantly improve performance and achieve high accuracy for downstream tasks which requires specialized knowledge. However, fine-tuning poses several challenges such as catastrophic forgetting \cite{kirkpatrick2017overcoming}. 
Minimizing the loss for adjusting the pre-trained model to a new task or data, could substantially drop the performance on some of the original seen data. 
    
RLHF (Reinforcement Learning from Human Feedback) is a broader adaptability technique compared to traditional fine-tuning that is only optimizing the model for limited tasks and domains. RLHF is a multifaceted training approach that refines the model behavior with human preferences and feedback~\cite{bai2022training}. RLAIF (Reinforcement Learning from AI Feedback) is similar to RLHF, however, it leverages another AI model to automatically generate feedback on the outputs of the base model being trained~\cite{lee2023rlaif}. Note that although these models could be aligned with some human values and feedback, it can also inherit the biases of the people whose feedback is involved in the model after fine-tuning.


\subsubsection{Prompt Engineering}
The training process for fine-tuning can be computationally expensive. Hence, instead of extensive fine-tuning, recent and popular paradigm of adaptation of foundation models is though well-crafted prompts that can guide foundation models to generate desired outputs or perform specific tasks without additional training. Popular prompt engineering and in context-learning approaches suggest that instead of extensive fine-tuning, well-crafted prompts can guide foundation models to generate desired outputs or perform specific tasks without additional training.

 
Although recent in-context learning efforts attempt to de-bias foundation models~\cite{dwivedi2023breaking}, these approaches have limitations. Since in-context learning does not modify model parameters, which would fundamentally increase the model's representational capacity, post-processing techniques alone cannot fully address representation disparity or alter the model's core understanding of marginalized communities. While in-context learning can influence model behavior through text conditioning, and safeguards can mitigate specific biases or manage harmful responses, these strategies offer limited bias reduction and cannot fundamentally change the model's inherent representations issues.


\section{Technical Gaps \& Call To Actions}
\label{sec:call_for_action}
\begin{figure*}
    \centering
    \subfigure[]{\includegraphics[width=0.18\textwidth]{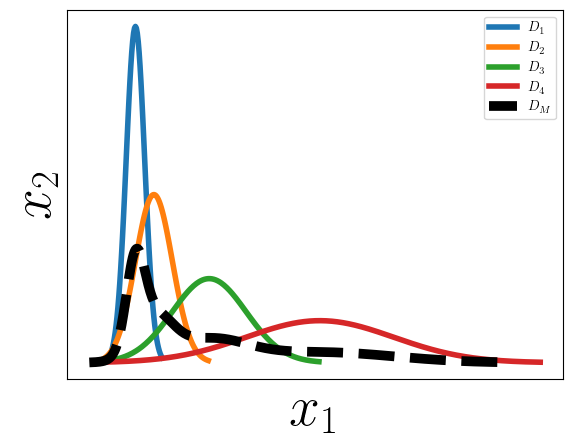}} 
    \subfigure[]{\includegraphics[width=0.18\textwidth]{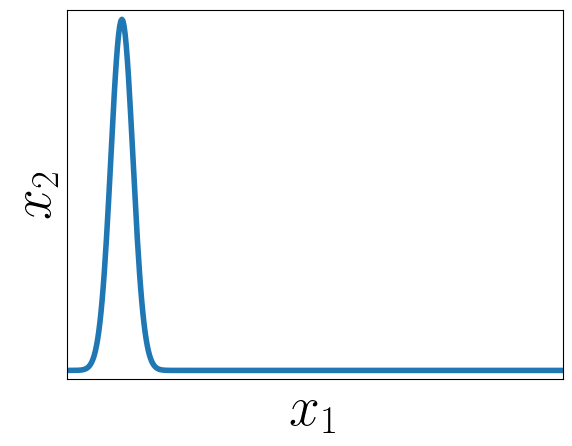}} 
    \subfigure[]{\includegraphics[width=0.18\textwidth]{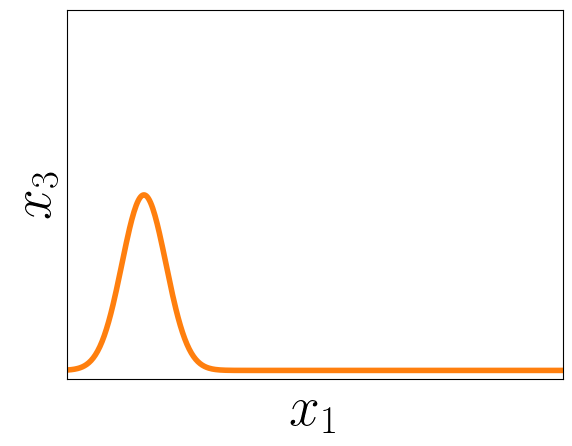}}
    \subfigure[]{\includegraphics[width=0.18\textwidth]{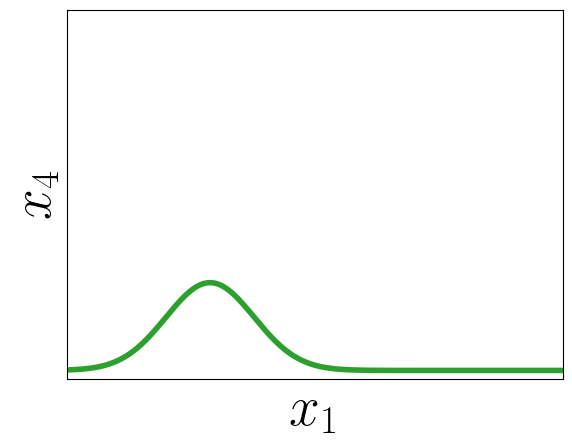}}
    \subfigure[]{\includegraphics[width=0.18\textwidth]{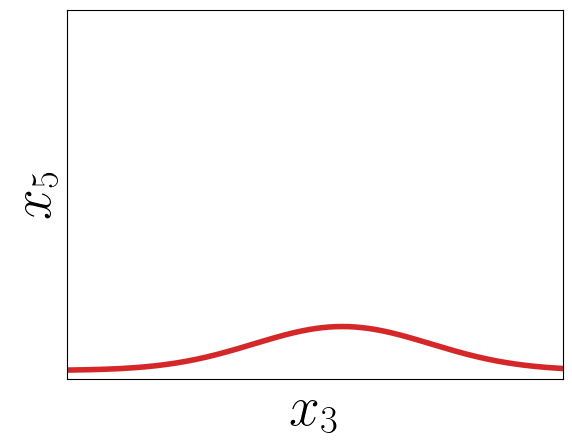}}
    \caption{
Simplified example of data with a mixture of heterogeneous distributions. The aggregated distribution, commonly assumed for training machine learning models, is represented in Figure (a) by a black dotted line $D_m$, while the underlying distributions are depicted in blue $D_1$, orange $D_2$, green $D_3$, and red $D_4$. In this example, all distributions are assumed to be Gaussian with equal weight, which is not reflective of real-world scenarios where marginalized communities often have smaller data sizes. Despite this simplification, the aggregated distribution still differs significantly from all the underlying distributions. Learning based solely on the aggregated distribution not only fails to accurately represent any of the underlying distributions, as shown in (a), but also risks missing variations if the dimensions of the underlying distributions differ, as shown in (b-e). While simply reweighting or adding more data points to marginalized distributions is not helpful, the complexities of the distributions can significantly impact the learning process and should be reflected in the method of aggregation.
}
    \label{fig:distributions}
\end{figure*}

As outlined earlier, marginalized communities experience multifaceted disparities rooted in the lifecycle of foundation models. Here, we mainly focus on the technical limitations that hinder efforts to address these issues. We will demonstrate how training methodologies, despite relying on high-quality data, can reinforce cascading disparity phenomena and we propose a set of call for actions to mitigate the root of such disparities, i.e., representation disparity. We acknowledge that the calls for action in this position paper are mainly technical, and we note other sociotechnical dimensions of disparities that are important to notice but out of the scope of our position paper in the impact statements.

\noindent \textbf{Call to Action: Developing Guidelines for Training under Mixtures of 
Heterogeneous Distributions}: First, we must emphasize that the data used to train foundation models often originates from diverse underlying distributions. One of the fallacies in training foundation models is the simplifying assumption that the underlying distribution is a long-tailed distribution (see Figure~\ref{fig:distributions}, sub-figure (a), $D_m$). While long-tailed distributions (like those in recommender systems due to popularity bias) are a relevant factor, specifically in traditional ML models, the disparities facing marginalized communities stem not just from limited data points, but from fundamental differences within their distributions. 

As we defined in Section~\ref{sec:marginalized}, we account for three characteristics for marginalized community, size, distinct distribution and distribution complexity. Hence, to mitigate representation disparity, this distinction with long-tailed distribution and the consideration of mixture of distribution is essential to grasp (see Figure~\ref{fig:distributions}, sub-figure (a)). Furthermore, even considering the mixture of distributions assumption, methods are needed to address high-dimensional datasets that inherently exist across multiple low-dimensional manifolds (see the left plot on Figure~\ref{fig:distributions}). We argue that considering low-dimensional manifolds to learn representation of each distinct distribution can significantly improve the representation of marginalized communities and reduce the representation gap. One could consider hierarchical manifold learning similar to hierarchical Bayesian models to capture global and local variances or dependencies, such as Bayesian meta learning~\cite{ravi2018amortized}.

While increasing model capacity could theoretically help learn under mixtures of heterogeneous distributions, it is important to consider practical limitations (i.e., learning all the underlying dimensions $x_1,x_2,x_3,x_4,x_5$ in Figure~\ref{fig:distributions}). The computational complexity of blindly pouring data into the training would quickly drive up costs, making it an unsustainable solution in real world applications. Here, we speculate less computational and resource extensive solutions to address the representation disparity by fixing the data ordering and batch aggregation problems discussed in section~\ref{sec:disparity_source}. We suggest to strategically organizing data points based on their underlying distributions. Grouping data from similar distributions minimizes conflicting gradient signals in training. We believe that success of foundation models to enhance downstream algorithmic fairness during adaptation phases by allowing changes to better reflect marginalized communities, that have been documented by existing work~\cite{mao2023last}, are due to better arrangements of data at such smaller scale. However, accurately grouping data requires a comprehensive understanding of underlying distributions.

The challenge of training machine learning models under heterogeneous data distributions is a significant area of research, particularly within the domain of federated learning~(FL). In FL, models are trained collaboratively across clients that possess diverse datasets, reflecting real-world scenarios where data is not uniformly distributed~\cite{mcmahan2017communication}. While conventional FL treats each client as a unique distribution, there are often underlying sub-distributions within the broader heterogeneous dataset. To address this, advanced clustering techniques has been employed to identify and group these distinct sub-distributions, enabling more targeted model training~\cite{ghosh2019robust,malekmohammadi2024mitigating}.

An additional challenge lies in distinguishing low-quality data distributions from those representing marginalized communities. There is no universal definition of low-quality data; however, techniques designed to mitigate data poisoning and adversarial attacks can inadvertently misclassify data from marginalized communities as low-quality. While channeling inferences based on data distribution may help address poisoning and adversarial attacks, extensive research is needed to differentiate between truly low-quality data and unique characteristics of marginalized groups' data distributions. 

\noindent \textbf{Call to Action: Metrics for Representation Disparity via Manifold Embedding}: Our second call to action, is a call to address the potential flattening of latent dimensions learned during training for marginalized communities. The flattening of latent dimensions suggests that the model may not fully capture the complexities of their distributions, unlike those of majority groups.

To better grasp this, consider Figure~\ref{fig:distributions}. If distribution $D_4$ (sub-figure (e)) relies primarily on dimensions $x_3$ and $x_5$, learning dimensions $x_1,x_2,x_3,x_4$ from other distributions might obscure crucial nuances of $x_5$.  While distributions can share common elements such as sentence structure, alphabet, or cultural norms in language, they also possess unique dimensions, e.g., consider $x_3$ in Figure~~\ref{fig:distributions} which is a shared dimension between distributions $D_2$ and $D_4$. Dedicating model capacity specifically to learn these distinct dimensions would enable far more accurate representation.

Based on the theoretical foundation of manifold embeddings~\cite{melas2020mathematical}, we need to build fair representation for marginalized communities within data~\cite{wan2021fairness}. Comprehensive metrics to measure the multifaceted impact of data selection bias on model behavior are lacking, hindering effective evaluation. Existing fairness benchmarks often lack robustness, present wide range of ambiguities, social science pitfalls~\cite{blodgett2021stereotyping,gallegos2023bias} and fail to address core issues within learned representations.
Also, there is need for metrics that go beyond measuring disparities in downstream tasks. 
In single task machine learning models, disparities are typically measured within the output space specific to that particular downstream task.
Consequently, most available metrics that consider fairness and disparities among different groups focus on evaluations within this output space. Foundation models, however, diverge from this paradigm as they are intended to serve as versatile representations applicable to a wide array of tasks, including those that may emerge in the future. Given the expansive scope of application for foundation models, it becomes imperative to consider metrics that are robust enough to assess group disparities within the representation space. Evaluating the learned manifold involves assessing various geometric properties that can capture richness and nuances of the data representation. We want to emphasize that simple distance metrics in embedding spaces, as proposed in the literature~\cite{zhao2017men}, fail to fully assess the quality of embeddings for different communities. Collapsing many dimensions during the embedding process leads to a loss of nuance for marginalized communities, potentially assigning identical embeddings to distinct concepts. Lipschitz continuity can be used to evaluate how smoothly the manifold transitions between different regions.\citet{szegedy2013intriguing} showed that for CNNs, a lower Lipschitz constant indicates more robust features and results in higher generalizability.



\noindent \textbf{Call to Action: Guiding Capacity via Mixture of Expert Models}: Our next call to action lies in developing measures that both quantify differences between distributions within the embedding space and determine the capacity a model should allocate to each distribution. 

Mixture of Experts~(MoE) models offer a promising approach to address the challenges inherent in training with heterogeneous data~\cite{fedus2022switch}. Their architecture, consisting of specialized "expert" neural networks and a learned "gating" network, enables intelligent routing of inputs to the most appropriate expert(s). This allows for targeted training on specific data aspects or sub-distributions, making MoEs well-suited to the nuanced complexities we've outlined.

Furthermore, recent research into sparse networks suggests that a substantial portion of model capacity may be dispensable without sacrificing accuracy on majority-class data. This opens up potential avenues for exploration using techniques like Parameter-Efficient Tuning (PEFT)~\cite{gordon2023morphing} or LoRA~\cite{hu2021lora} to optimize MoE performance and reduce computational overhead.

In addition, PEFT has a potential to facilitate collaborative federated model training with low-resource marginalized communities. By tailoring resource allocation based on the size of individual communities distribution, we can ensure privacy protection in federated paradigms while avoiding the mismatched noise levels that often arise in differential privacy applications sensitive to distribution size (see Section~\ref{sec:disparity_type}).



\noindent \textbf{Call for Action: Model-guided Data Collection}: Our last call for action is to address a fundamental challenge faced by marginalized communities which is the non-uniform sampling of their data (see Section~\ref{sec:disparity_source}). This leads to frequent distribution shifts (as discussed in Section~\ref{sec:disparity_type}) and hinders model generalization. To counter this, we propose automated methods to identify sparse areas within marginalized data distributions and strategically employ active learning techniques~\cite{wang2017active}. Active learning enables models to proactively query for the most informative data points, guiding targeted curation and collection efforts to improve model performance on underrepresented areas~\cite{holzinger2016interactive}.

Recent fine-tuning techniques like RLHF and RLAIF (see Section~\ref{sec:disparity_source}) can be adapted to guide active data creation mechanisms, too. This offers a targeted approach to alleviate data sparsity within marginalized communities~\cite{hemmat2023feedback}. Our proposal emphasizes strategic data acquisition over simplistic model scaling or unguided data curation. While existing model capacity might be sufficient to learn marginalized data distributions, focused data collection is crucial. However, it's important to note that solely relying on model-identified sparsity may not fully reflect real-world data complexities. This approach primarily serves to improve the efficiency of data collection efforts. We fully acknowledge the potential for model-guided data collection to perpetuate existing biases. We emphasize that our approach is not a replacement for addressing fundamental data quality issues like the lack of digitized data in low-resource languages or marginalized communities. However, we believe that uncritically collecting more data can also reinforce existing disparities. Our method intentionally seeks out underrepresented areas, aiming to break those cycles. Our goal in this call for action is to optimize the data collection process within existing constraints. By identifying areas where the model's representations are inadequate, we can target data collection to fill gaps and broaden the model's understanding of underrepresented distributions instead of blindly increasing the dataset size. The inspiration from active learning highlights this point – the model itself suggests where to focus collection efforts. We propose a dynamic process where the model actively identifies its representational shortcomings. This targeted approach could potentially be more effective in broadening a model's understanding, especially in multilingual settings where conceptual gaps are readily detectable e.g., the significant variation, or even near-zero distance, between representations of distinct concepts across or within languages.

Finally, we believe covering all the nuances of the representation disparity and how all other courses of disparity stemmed from it by studying it thought the ML pipeline is an important aspect to picture and show the complexity of the cascading phenomena. Our focus on addressing this root cause through training procedures and novel metrics is an intentional and focused call to action. Current metrics are insufficient, and we believe developing new evaluations and metrics targeting capacity and embedding gaps is crucial for tackling representation disparities. Furthermore, as discussed earlier, we believe overemphasizing on solutions such as scaling laws and indiscriminate data collection can mask biases favoring majority groups. Recent literature on pruning indicates that we may be using unnecessarily large models for majority representation and it is even an *overkill”. In this position paper, we advocate for optimizing model capacity during training to ensure inclusivity across diverse distributions, i.e., MoE models. We emphasize that our calls to action are intentionally interconnected, forming a unified framework to address the cascading effect and prioritizing one call to action over another would miss the systemic nature of the problem.

\section*{Acknowledgments}
We would like to thank Hugo Larochelle for insightful discussions at early stages of the project, and Stephen Pfohl for his valuable feedback on later drafts. We also thank our reviewers for their insightful comments and constructive feedback.

\section*{Impact Statement}

Addressing the multifaceted problem of disparities in foundation models requires a holistic sociotechnical approach that goes beyond isolated technical fixes. While in previous section, we solely focused on technical gaps and position our work around the cascading disparity towards better training paradigms, we would like to emphasize on the need for \textbf{community engagement} and involve marginalized communities throughout the development process to ensure that their perspectives and concerns are addressed.

Foundation models democratize AI by empowering smaller organizations and communities with limited data or expertise. They unlock previously impossible applications like creative text generation and code translation. However, they also introduce unique challenges that emerge after deployment which impact marginalized community. While a comprehensive discussion of the societal impacts of foundation models is beyond the scope of this work, to conclude our paper, next, we highlight three critical areas that directly affect marginalized communities and require urgent attention.

\noindent\textbf{Measurements beyond Representation Disparity}: The size and complexity of foundation models make it difficult to comprehensively assess their disparities. Traditional disparity detection methods often fall short due to the model's ability to mask disparities in subtle ways. Foundation models, specifically multi-modal versions, often process multiple types of data, such as text, images, and audio. This means that disparities can manifest in different forms, making it challenging to develop a universal evaluation framework. Additionally, disparities can emerge in different contexts and evolve over time. Static evaluation methods may not capture these dynamic patterns. Finally, currently, no widely accepted set of metrics exists to measure disparities in foundation models. Different researchers and organizations may prioritize different fairness criteria, which makes comparisons and bench marking very difficult and challenging.

\noindent\textbf{Monopoly Effects}: Unlike traditional ML models, training foundation models necessitates vast amounts of data and specialized hardware, often inaccessible to most organizations and researchers. This creates a high barrier to entry for marginalized communities, and leads to a concentration of power among a few large tech companies with the resources to develop and control these models. This concentration can stifle competition and innovation, as smaller players are often unable to challenge the dominance of large players with their proprietary foundation models. Moreover, as more organizations rely on foundation models from a few providers to power their products and services, a widespread dependence on these models emerges which lead to propagation of disparities in various downstream tasks that are stemmed from a single model. This can also create a lock-in effect, which makes it difficult to transition to alternative models or providers, as entire ecosystems become built upon a limited set of foundation models. This concentration of power grants a few companies significant control over the development and direction of AI research and applications while the root of disparities are the same and create a systematic way that disparities impact society, specifically, marginalized communities. 

\noindent\textbf{Long-term Impact}: Foundation models undergo training using data obtained from internet scraping. As mentioned in previous sections, this data is not selectively chosen; rather, it is randomly sampled from the internet. Consequently, the distribution of this data is biased, reflecting the cultures and attributes of regions with greater internet access and usage. Given that foundation models find application in numerous generative tasks across various modalities (e.g., text, image, video, music, etc.), this exacerbates the existing gap in the representations of data on the internet. The data used to train these foundation models in subsequent iterations contributes to a snowball effect, potentially resulting in almost zero representations of certain marginalized groups.

\noindent\textbf{Evaluation for downstream task}: 
Addressing various biases towards marginalized communities in foundation models requires an in-depth understanding of interconnected harms and the development of holistic solutions. Future research should focus on the development of robust metrics and approaches that surpass surface-level evaluations. Existing work suggests conflicting empirical results about the relations of intrinsic biases and extrinsic biases~\cite{gupta2022how,kiela2022evaluating,stafanovics2022mitigating,mohanty2022do}. We encourage future research to focus on theoretical foundations to analyze the relations between various intrinsic biases and extrinsic biases through the lens of representation disparity.


\bibliography{bibtex}
\bibliographystyle{icml2024}

\end{document}